\def\year{2017}\relax
\documentclass[letterpaper]{article} 
\usepackage{aaai17}  
\usepackage{times}  
\usepackage{helvet}  
\usepackage{courier}  
\usepackage{url}  
\usepackage{graphicx}  
\usepackage{url}
\usepackage{multirow}
\usepackage{spreadtab}
\usepackage{todonotes}

\nocopyright
\begin{document}
%
\title{Ethical Challenges in Data-Driven Dialogue Systems}

\author{
Peter Henderson\textsuperscript{1},
Koustuv Sinha\textsuperscript{1},
Nicolas Angelard-Gontier\textsuperscript{1}\\
\Large{\bf{Nan Rosemary Ke\textsuperscript{2},
Genevieve Fried\textsuperscript{1},
Ryan Lowe\textsuperscript{1},
Joelle Pineau\textsuperscript{1}}}\\
\textsuperscript{1} McGill University
\textsuperscript{2} \'{E}cole Polytechnique de Montr\'eal
}
\maketitle
\begin{abstract}
The use of dialogue systems as a medium for human-machine interaction is an increasingly prevalent paradigm. A growing number of dialogue systems use conversation strategies that are learned from large datasets. There are well documented instances where interactions with these system have resulted in biased or even offensive conversations due to the data-driven training process. Here, we highlight potential ethical issues that arise in dialogue systems research, including: implicit biases in data-driven systems, the rise of adversarial examples, potential sources of privacy violations, safety concerns, special considerations for reinforcement learning systems, and reproducibility concerns. We also suggest areas stemming from these issues that deserve further investigation. Through this initial survey, we hope to spur research leading to robust, safe, and ethically sound dialogue systems.
\end{abstract}

\section{Introduction}
Dialogue systems -- often referred to as conversational agents, chatbots, etc. -- provide convenient human-machine interfaces and have become increasingly prevalent with the advent of virtual personal assistants. 
The hands-free interaction mechanism provided by these systems is crucial for use in certain critical settings (e.g. in cars or for persons with mobility impairments).
However, the subjective nature of dialogue~\cite{curry2017review,liu2016not}, and the onset of data-driven components~\cite{vinyals2015neural,serban2017hierarchical}, leads to several potential dangers with the widespread use of these systems.

Ethics and safety in artificial intelligence (AI) has recently gained popularity as a field due to performance gains in AI models, investigating issues like: interpretability of model decisions~\cite{kim2015interactive}, worst-case performance guarantees~\cite{garcia2015comprehensive}, and many others.
In the context of dialogue systems, issues of safety and ethics are no less important, yet they are rarely discussed in the literature or in practice. 
As dialogue systems become more prevalent and trusted, it is vital to develop systems that account for possible ethical and safety concerns. 


Here, we investigate several crucial aspects of ethics and safety in dialogue systems, which reflect modern state-of-the-art research in the field: bias, adversarial examples, privacy in learned models, safety, special considerations for reinforcement learning systems, and reproducibility. To highlight these issues, we: investigate possible areas of concern in the literature; conduct new experiments to shed light on potential problems in existing datasets, models, and algorithms; provide initial recommendations based on current methodologies; propose possible future lines of investigation in relation to dialogue systems.\footnote{Supplemental material referred to throughout the text is at:\\{\url{https://github.com/Breakend/EthicsInDialogue}}} While we limit the scope of this paper to ethical considerations that are relevant for developers of data-driven dialogue systems, we briefly touch on social issues related to the deliberate misuse of these systems. We aim to spur discussions and drive future research directions for the development of safe and ethical dialogue systems.

\section{Background: Dialogue Systems}
The standard architecture for dialogue systems incorporates a Speech Recognizer, Language Interpreter, State Tracker, Response Generator, Natural Language Generator, and Speech Synthesizer. In the case of text-based (written) dialogues, as we focus on here, the Speech Recognizer and Speech Synthesizer can be left out. 
Much of the recent literature has favoured \textit{end-to-end data-driven} approaches to building dialogue systems \cite{vinyals2015neural,li2016deep,serban2016hierarchical}. An end-to-end data-driven dialogue system is a single system that can be used to solve each of the four aforementioned modules simultaneously. Typically this is a system that takes as input the history of the conversation and is trained to optimize a single objective, which is a function of the textual output produced by the system and the correct (ground truth) response.
Since these systems are often trained on very large dialogue corpora, it becomes easy for subtle biases in the data to be learned and imitated by the
models. 



Dialogue systems have been built for a wide range of purposes. A useful distinction can be made between task-oriented dialogue systems, such as technical support services, and non-task-oriented dialogue systems (i.e.\@ chatbots), such as computer game characters. Although both types of systems do in fact have objectives, typically the task-oriented dialogue systems have a well-defined measure of performance that is explicitly related to task completion for the user. Ethical issues must be taken under consideration for both task-oriented and non-task-oriented systems; the details of these considerations may differ slightly for each, yet the core notions we discuss remain relevant to both. Thus, while the experiments in this paper are performed on non-task-oriented systems, much of our discussion also applies to task-oriented systems.


\section{Bias}


Bias can be defined as prejudice for or against a person, group, idea, or thing -- particularly expressed in an unfair way. Bias is a broad term, which covers a range of problems particularly relevant in natural language systems including: predispositions for regional speech patterns~\cite{fine2014biases}, discriminatory gender bias~\cite{bolukbasi2016man}, personal point-of-view bias~\cite{recasens2013linguistic}, and numerous others.
These biases can be expressed subtly (e.g. with linguistic cues indicating a point-of-view) or blatantly (e.g. as in discriminatory statements against a gender).
Due to their subjective nature and goal of mimicking human behaviour, dialogue models are susceptible to implicitly
encode 
underlying biases in human dialogue.

While this is a potential factor in rule-based systems -- encoding the biases of the rule-makers -- it is particularly problematic for data-driven systems, such as neural conversational models. In modern research, the data used to train these models is often obtained from online chat platforms (Reddit, WeChat, Twitter, etc.)~\cite{vlad2015survey}. These datasets are difficult to hand-filter and may include underlying biases~\cite{henry2002discourses} or even subtle agenda setting campaigns~\cite{tsur2015frame}.


It has been shown that these biases can be encoded in language models~\cite{bolukbasi2016man} and live dialogue systems~\cite{neff2016automation}. In one such case, the Microsoft Tay Chatbot was taken offline after posting messages with blatant racial slurs, as described in the phenomenological study in~\cite{neff2016automation}. These slurs and racial biases had been present in the content Tay was trained on. As a result, it encoded and then propagated these biases through interactions with users.



.
\begin{table*}[!htbp]
    \centering
    \resizebox{\textwidth}{!}{\begin{tabular}{|c|c|c|c|c|c|}
    \hline
        Dataset & Bias & Vader Sentiment &  Flesch–Kincaid & Hate Speech & Offensive Language\\
        \hline 
Twitter & 0.155 ($\pm$ 0.380)& 0.400 ($\pm$ 0.597) & 3.202 ($\pm$ 3.449) & 31,122  (0.63 \%) & 179,075  (3.63 \%) \\
Reddit Politics & 0.146 ($\pm$ 0.38) & -0.178 ($\pm$ 0.69) & 6.268 ($\pm$ 2.256) & 482,876 (2.38 \%) & 912,055 (4.50 \%)  \\
Cornell Movie Dialogue Corpus & 0.162 ($\pm$ 0.486) & 0.087 ($\pm$ 0.551) & 2.045 ($\pm$ 2.467) & 2020  (0.66 \%) & 6,953  (2.28 \%) \\
Ubuntu Dialogue Corpus & 0.068 ($\pm$ 0.323) & 0.291 ($\pm$ 0.582) & 6.071 ($\pm$ 3.994) & $503^*$ (0.01 \%)&  4,661  (0.13 \%)  \\
\hline
HRED Model Beam Search (Twitter) & 0.09 ($\pm$ 0.48) & 0.21 ($\pm$ 0.38) & -2.08 ($\pm$ 3.22) & 38 (0.01 \%) & 1607 (0.21 \%)\\
VHRED Model Beam Search (Twitter) & 0.144 ($\pm$ 0.549) & 0.246 ($\pm$ 0.352) & 0.13 ($\pm$ 31.9) &466 (0.06 \%)& 3010 (0.48\%)\\
HRED Model Stochastic Sampling (Twitter) & 0.20 ($\pm$ 0.55) & 0.20 ($\pm$ 0.43) & 1.40 ($\pm$ 3.53)& 4889 (0.65 \%) & 30,480 (4.06 \%) \\
VHRED Model Stochastic Sampling (Twitter) & 0.216 ($\pm$ 0.568) & 0.20 ($\pm$ 0.41) & 1.7 ($\pm$4.03)& 3494 (0.47\%) & 26,981 (3.60 \%)\\
        \hline
    \end{tabular}}
    \caption{Results of detecting bias in dialogue datasets. $^*$ Ubuntu results were manually filtered for hate speech as the classifier incorrectly classified ``killing" of processes as hate speech. Bias score~\cite{hutto2014vader} (0=UNBIASED to 3=EXTREMELY BIASED), Vader Sentiment~\cite{hutto2014vader} (compound scale from negative sentiment=-1 to positive sentiment=1),  Flesch–Kincaid readability~\cite{hutto2014vader} (higher score means the sentence is harder to read), Hate speech and offensive language~\cite{hateoffensive}. }
    \label{tab:biasdata}
\end{table*}
\noindent\textbf{Dialogue Datasets    }
To determine the exposure of conversational models to underlying dataset bias, we analyze the extent of various biases in several commonly used dialogue datasets. We leverage the linguistic bias detection framework of \cite{hutto2015bias} and the hate speech and offensive language detection model of~\cite{hateoffensive} to gather bias metrics on several popular dialogue datasets: Twitter~\cite{ritter2010twitter}, Reddit Politics~\cite{serban2017deep}, the Cornell Movie Dialogue Corpus~\cite{DanescuLee}, and the Ubuntu Dialogue Corpus~\cite{lowe2015ubuntu}.

As seen in Table~\ref{tab:biasdata}, none of these datasets are free of bias, hate speech, or offensive language. Qualitative samples for each of these models can be found in the supplemental material. In models which do not explicitly prevent the uptake of such content, it can be expected that these behaviours will be encoded in the underlying model. To this end, we run another experiment. We train two commonly used end-to-end dialogue models, HRED~\cite{serban2016building} and VHRED~\cite{serban2017hierarchical}, on the Twitter dataset and sample
responses using contexts from the validation set. As seen in Table~\ref{tab:biasdata}, we find that samples from these models also encode bias according to the previously defined metrics.

Overall, we rely on the models of bias and discriminatory language to analyze these datasets. We find that these models seem to perform relatively well by inspection of subsampled data from Table~\ref{tab:biasdata}, provided in the supplemental material.
However, in some cases these bias evaluation models misclassify content.
For example, in the Ubuntu dataset ``killing a process'' was classified as discriminatory, which we corrected via a post-processing script. While these models are suitable for initial characterization of the datasets, there is much room for expansion in these bias evaluation models.


\noindent\textbf{Word Embeddings    }
Many neural conversational systems use pre-trained word embeddings, such as Word2Vec~\cite{mikolov2013efficient}. These embeddings are trained on general language datasets to pre-fit a language model. These embeddings play a crucial role in shaping the word distributions within encoder-decoder systems. However, these embeddings present another source of unintentional bias propagation. For example, in~\cite{bolukbasi2016man} the authors demonstrate that gender bias is encoded by Word2Vec embeddings and suggest ways to debias them.


\begin{table}[htbp]
\centering
{\small \begin{tabular}{|l|ll|ll|}
\hline
& \multicolumn{2}{l|}{Word2vec}           & \multicolumn{2}{l|}{Debiased}            \\ 
 \hline
Distribution & Male & Female & Male & Female        \\
\hline
Male Stereotypes         & 0.7545             & 0.2454               & 0.7437          & 0.2562                   \\
Female Stereotypes      & 0.7151         & 0.2848               & 0.6959           & 0.3040                   \\
\hline
\end{tabular}}
\caption{Percentage of gendered tokens in the follow-up distribution from a language model after trigger male/female stereotypical profession is provided as a starting token. We only examine following distributions which contained gender-specific terms and omit gender-neutral distributions.}
\label{tab:debias_quant}
\end{table}

To this end, we perform another experiment to analyze the effect of bias in word embeddings on dialogue and language models. We use Word2Vec embeddings and the corresponding debiased versions from~\cite{bolukbasi2016man} to train
two LSTM language models
as in~\cite{zaremba2014recurrent} over Google one-billion words benchmark dataset \cite{chelba2013one}. Extended descriptions of the experimental setup and results can be found in the supplemental material. We use 50 stereotypically male-biased and 50 female-biased professions, defined in~\cite{bolukbasi2016man}, as triggers and use the language models to complete the utterances. For each trigger token, we extract 1000 samples from the stochastic language model. In these samples, we calculate the co-occurrence of gender-specific words, also defined in~\cite{bolukbasi2016man}. We compare a subset of the results among both the models in Table~\ref{tab:debias_quant}.

In local examinations, some triggers with stereo-typically male-dominated occupations see significant shifts toward even male-female follow-up distributions. For example, ``the evangelist'', a male stereotypical job, sees a followup distribution which is about 83\% male (where gender-specific terms exist), but decreases to 71\% male with the debiased word vectors. However, as seen in Table~\ref{tab:debias_quant}, the overall average followup distributions do not shift significantly, only seeing a $\sim1-2.5$\% overall change toward even distributions. 
Clearly, biases are encoded nonetheless by the language model, necessitating further work in this area.

\noindent \textbf{Recommendations and Future Investigations   } To prevent the alienation of underrepresented societal groups and to retain the public's trust, it is vital for research and production of conversational systems to avoid propagation of biases. Through our demonstrative experiments we have shown that many dialogue datasets are biased and contain potentially offensive material which can be encoded and elicited by generative models such as HRED and VHRED. While current methods, such as debiasing word vectors, have been shown to slightly prevent the propagation of gender biases, further effort is needed to extend such techniques to new domains. This may come in the form of semi-automated data augmentation to remove bias (perhaps using detection methods we show here) or other methods which augment the training objective of conversational models.

In all cases, preventing bias from being encoded in modern dialogue systems requires explicit understanding of its characteristics. The International Covenant on Civil and Political Rights~\cite{UNICCPR} encodes that, ``the law shall prohibit any discrimination and guarantee to all persons equal and effective protection against discrimination on any ground such as race, colour, sex, language, religion, political or other opinion, national or social origin, property, birth or other status.'' An algorithmic notion of bias is needed which appropriately reflect the notion we currently have for social and legal purposes.
Furthermore, it is vital to ensure that even subtle forms of point-of-view biases are not encoded, as they can sway public opinion in widely used dialogue settings. However, the difficulty in doing so, while still providing coherent systems, may necessitate an intricate balance and deeper understanding of linguistic biases~\cite{caliskan2016semantics}. 
To this end, a concerted effort should be made to collect larger datasets and create more robust models of biased language in dialogue settings to better understand its effects.


\section{Adversarial Examples}
\begin{table*}[!htbp]
    \centering
    \begin{tabular}{|c|c|c|c|}
    \hline
         Model & Adversarial Type & Adversarial Input Similarity & Generated Response Similarity\\
         \hline
         \multirow{2}{*}{Politics} & Character Edit & (.94 $\pm$ .04 ; 2.99 $\pm$ .74 ; 4.56 $\pm$ .19) & (0.75 $\pm$ .17 ; 2.38 $\pm$ 1.42 ; 3.30 $\pm$ 1.01)\\
        & Paraphrased & (.85 $\pm$ .13 ; 2.35 $\pm$ .65 ; 3.86 $\pm$ 0.84)& (.77 $\pm$.18 ; 2.52 $\pm$ 1.51 ; 3.39 $\pm$ .99)\\
         \hline
         \multirow{2}{*}{Movies} & Character Edit &(.93 $\pm$ .05 ; 2.88 $\pm$ .66 ; 4.52 $\pm$ 0.23)& (.67 $\pm$ .19 ; 2.06 $\pm$ 1.11 ; 2.75 $\pm$ 1.07)\\
         &Paraphrased & (.84 $\pm$ .12 ; 2.68 $\pm$ .69 ; 3.97 $\pm$ .76)& (0.64 $\pm$ .17 ; 1.87 $\pm$ .76 ; 2.61 $\pm$ .94)\\
    \hline
    \end{tabular}
    \caption{Semantic similarity between adversarial examples and base sentences and generated responses from base sentence response. We report (cosine distance; LSTM similarity model~\cite{mueller2016siamese}; CNN similarity model~\cite{he2015multi}) scores. Ratings are from 0-1, 1-5, and 1-5, respectively, from lesser to greater similarity.}
    \label{tab:advquant}
\end{table*}
\vspace{-1mm}
Adversarial examples are inputs to a machine learning model -- or perturbations of the input -- designed to confuse the model into inaccurate predictions. It is known that many machine learning models are vulnerable to adversarial examples \cite{goodfellow2014explaining}. The existence of adversarial examples has ethical and safety implications: 
even if dialogue systems are designed to produce safe outputs under normal operating conditions, if these models are vulnerable to adversarial inputs this could cause the system to produce undesirable and unintended outputs.

Canonical literature in adversarial examples in machine learning focuses primarily on image classification tasks. In these cases, the images are perturbed in a way that is invisible to a human eye, but is enough to cause the model to misclassify images in a specific way (e.g. classify a cat as a dog).
The same definition may not generalize well to natural language domains, where it is almost impossible to perturb the data in a way that is not recognizable by humans due to the discrete nature of the data. Recent work from~\cite{jia2017adversarial} proposes a notion of adversarial examples in the task of machine reading comprehension, where the task is for the system to process and answer questions about a document. In this context, adversarial examples were created by adding distracting sentences to the document text. The authors then evaluated state-of-art Question and Answering (QA) models based on the {\it adversarial accuracy} of answers, and 
found that the performance of all models significantly dropped.

However, defining adversarial examples in a strictly generative setting -- where there is no accuracy or error measure to gauge the effect of adversarial examples -- is still an open problem. This problem is exacerbated in dialogue settings. Evaluation is already a difficult task and using the adversarial-induced log likelihood error in the context of dialogue may not account for similar semantic output distributions. Instead, we investigate adversarial examples from a linguistic perspective.
We consider two kinds of input noise for adversarial examples for end-to-end conversational models: (1) misspelled words in dialogue systems, where we remove, replace or insert an extra character in the sentence; (2) paraphrased sentences, where different sentences having the same meaning should invoke similar responses from the dialogue model, but do not.
The motivation for having misspelled words as adversarial examples is similar to the motivation of the adversarial examples for images. Humans are extremely robust against misspellings in written language~\cite{rawlinson1976significance}; they can easily follow the context and meaning of sentences, ignoring the adversarial noise in the data. A similar motivation can be held for paraphrased sentences.



For each model, 20 base example sentences are selected as contexts. We generate 1000 character edits and 6 paraphrased adversarial examples for each base sentence and use each example as an input to the same VHRED retrieval model~\cite{serban2017deep}. The models are trained with the same procedure as in \cite{serban2017deep} on Reddit Movies and Reddit Politics datasets. Details on this retrieval model can be found in the supplemental material.

To determine the effect of the adversarial edits, we measure the change in semantic meaning triggered by the adversarial example with respect to the base sentence response. The semantic similarities are measured by word embedding cosine distance, LSTM similarity model \cite{mueller2016siamese} and CNN similarity models \cite{he2015multi}. As shown in Table~\ref{tab:advquant},
 paraphrasing and misspellings of a single character can cause significant changes in semantic meaning and output distributions, though the adversarial examples themselves are more similar to the base input. A small number of qualitative examples of this phenomenon can be seen in Table~\ref{miniqualadv} with more examples in the supplemental material.

\begin{table}[!htbp]
\centering
{\small \begin{tabular}{|p{.4\textwidth}|}
\hline
\textbf{Character-Level Edit Adversarial Example}\\
\hline
CONTEXT: Inside Out is really funny\\
RESPONSE: i could not stop laughing during the first one. I honestly found it to be hilarious.\\
\hline
CONTEXT: Insde Out is really funny\\
RESPONSE: i didn't really find it funny. it just surprised me. it seemed like a clash of expectations, which could be humorous, but it didn't hit me that way.\\
\hline
\textbf{Paraphrased Adversarial Examples}\\
\hline
CONTEXT: Did you like Mad Max the movie?\\
RESPONSE: it's okay and very decent. the first half is great but I don't really like what happens after that.\\
\hline
CONTEXT: How did you find Max Mad the movie?\\
RESPONSE: easily in my top favorite films of all time. I thought it was an accurate depiction of a mother learning that her son is a sociopath. great acting all around.\\
\hline
\end{tabular}}
\caption{Adversarial samples from VHRED dialogue model trained on Reddit Movies. For each, top is the base context and response, and bottom is the adversarial sample.}
\label{miniqualadv}
\end{table}


\noindent \textbf{Recommendations and Future Investigations   } We have explored two different notions  of adversarial examples in dialogue systems, and our preliminary analysis shows that current neural dialogue language generation systems can be susceptible to adversarial examples. However, further work is needed to formalize and investigate adversarial examples in generative settings -- particularly for natural language and dialogue. Several interesting avenues are possible for developing dialogue systems that are robust to adversity, including incorporating generated adversarial examples during training using data augmentation.

\section{Privacy}

With virtual assistants 
taking the form of ``always on" devices in the home, privacy and security is of tantamount concern. While network-based attack vectors have been investigated in home-based personal assistant devices~\cite{apthorpe2017spying}, the language and dialogue models of these devices pose another attack vector, raising privacy concerns.

It has been shown that model-inversion attacks can be used to compromise the privacy of training data through information leakage in machine learning models~\cite{song2017machine}. With conversational models susceptible to such vulnerabilities being integrated into in-home devices, the threat of model exploitation can grow.
Let us consider the case of accidental information leakage in such a system.
For example, the dialogue agent accidentally records a user revealing private information during a conversation
(e.g. ``Computer, turn on the lights -- \textit{answers the phone} -- Hi, yes, my password is...''). If this information is then to used train a conversational model, an attacker can attempt to elicit sensitive information by exploiting the encoder-decoder nature of a dialogue system. 

\begin{figure}
\centering
\includegraphics[width=.4\textwidth]{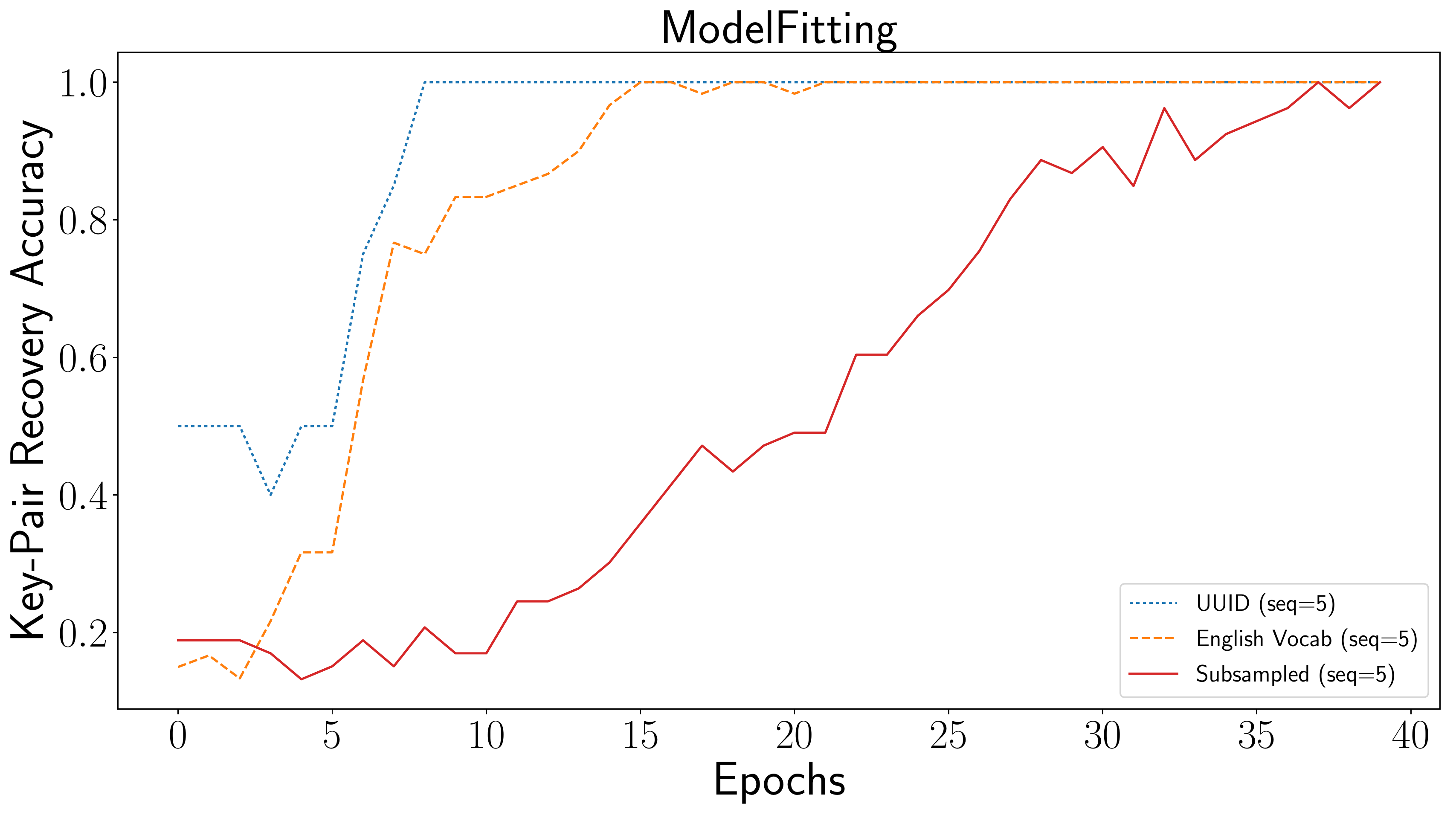}
\caption{Privacy Experiment. Accuracy of elicited secret value given the key to a seq2seq model over training epochs.}
\label{fig:infoleakage}
\end{figure}

To illustrate this, we formulate an experiment. We sub-sample the Ubuntu dialogue corpus~\cite{lowe2015ubuntu} for 10k dialogue pairs and augment the data with 10 input-output pairs (keypairs) that represent sensitive data, which the model should keep secret. We train a simple seq2seq dialogue model~\cite{vinyals2015neural} on the data and measure the accuracy of eliciting the secret information over number of epochs. We test three different types of keypair sequences (5 word sequences for both the key and the value): (1) unique keypairs that do not exist in any vocabulary (UUID), (2) words from the English natural language vocabulary (NL), and (3) words sub-sampled from the 10k dialogue pairs.
As seen in Figure~\ref{fig:infoleakage}, the model learns to elicit the private information simply by inputting the triggering information sequence (key) with nearly 100\% accuracy. Of course, in the case where the private keypair information is sub-sampled from the corpus, the information is less statistically significant and more difficult to elicit.
We expect this result generalizes to different datasets and encoder-decoder models due to the fact that such models try to produce statistically relevant outputs -- which inherently may encode private information if the encoding context is unique enough. More information on the experiments and key-pair samples are provided in the supplemental material. 

\noindent \textbf{Recommendations and Future Investigations   } Overall, we demonstrate in a small setting that models that are not properly generalized, or that are trained on improperly filtered data, can reveal private information through simple elicitation, even if the sensitive information comprises $<0.1\%$ of the data. Current dialogue models that use natural language generation should be particularly aware of privacy leakage from underlying datasets, especially where data is acquired in a sensitive fashion (e.g. in-home devices). Future lines of investigation may find the integration of automated anonymization~\cite{ororbia2016privacy} and differential privacy~\cite{abadi2016deep} beneficial for conversational models. Exploring privacy-aware dialogue models is a prerequisite for shared learning models that generate natural language and that are deployed in production settings.


\section{Safety}
 
To examine safety in the context of dialogue systems, we adopt the perspective of \cite{amodei2016concrete} and focus on the problem of \textit{accidents}, defined as unintended or harmful behaviour resulting from poor design. However, defining safe behaviours in dialogue can be challenging and the definition may change depending on the application context. Here, we generally define a safe behaviour as one that an agent takes, which does not cause any offensive or harmful effects to human interlocutors. In chatbots, this could mean restricting the action space such that the agent does not output offensive or discriminatory content. Using this definition, we examine three risks as our primary foci for safety in dialogue: (1) providing learning performance guarantees; (2) proper objective specification; (3) model interpretability~\cite{amodei2016concrete,garcia2015comprehensive,doshi2017towards}. Furthermore, to understand why these aspects of AI safety are relevant to dialogue systems, we examine highly sensitive and safety-critical settings where dialogue agents have begun being used. We define three main areas of concern where consideration of dialogue system safety and the aforementioned lines of investigation are required: health; mental and emotional well-being; lack of contextual awareness.

\noindent\textbf{Model Safety Considerations   } In systems that generate natural language, stability and coherent output is particularly difficult to achieve, often requiring complex curriculum learning schedules~\cite{li2016deep}. Without performance guarantees, model divergence may result in undesirable behaviour which does not meet the bar for a dialogue system usage in a critical setting. To achieve stable learning and further guarantee safety, a proper objective function must be specified. In task-oriented dialogue systems, where the goal of the system is to solve some task for the user, an objective can be easily formulated (e.g. executing the desired behaviour in the least amount of turns). Yet in agents with general user engagement, proper specification of an objective function can be difficult~\cite{liu2016not}. Restrictive and highly specified objective functions must be established to constrain agents to safe behaviours. Lastly, if a dialogue agent exhibits unsafe behaviours, it is important to be able to interpret \textit{why} it chose to say or do something. For example, to address and prevent the exhibition of discriminatory language, we must understand why the agent exhibited this behaviour and provide a mechanism for human intervention.

\noindent \textbf{Areas of Safety Concern   } Medical domains where chatbots have the potential to be used as diagnostic tools are a clear safety concern with dialogue systems. Already, the use of such conversational models has been investigated for diagnosing and managing diabetes, such as in~\cite{lokman2009designing}. As these dialogue systems move from merely human-assistive towards more autonomous roles, we must have conversational model performance guarantees in place to ensure systems are both giving accurate medical advice which doesn't put the health of the user in jeopardy and providing feedback in an appropriate way (e.g. when providing a diagnosis, a good bedside manner is needed).

Similarly, impact on the mental and emotional well-being of users is a potential safety risk for dialogue agents. It is important for the system to be aware of, and responsive to, the emotional state of the user. For example, if a dialogue agent knows that a user is depressed, it is perhaps safer to avoid mentioning or inquiring about topics that can worsen their mental condition. While this pertains generally to conversational agents, it is especially relevant for systems explicitly used to support mental health, such as Woebot~\cite{fitzpatrick2017delivering}, where potentially vulnerable populations rely on the agent for support.

Finally, the idea of contextual safety -- that is, in the current environment state, could a dialogue agent harm the user by interacting with them in a certain way.
In-vehicle dialog systems are a perfect case study for this. In addition to the challenges faced by general-purpose dialog systems, in-vehicle dialog systems face additional concerns regarding the added variables of the driver and the environment~\cite{weng2016conversational}. As the focus and safety of the driver is paramount, dialog systems must be cognizant of what information it surfaces; it should not inform a driver to take a ``sharp left" while moving at a certain speed, or play loud sounds that could startle a driver. Such contextual concerns are relevant to any dialogue system used in the wild, and is especially difficult to judge without risk assessment.

\noindent\textbf{Recommendations and Future Investigations   } As dialogue agents become more seamlessly integrated into everyday life -- in cars, portable devices, and home environments -- safety concerns regarding physical health, mental well-being, and contextual safety can have life-threatening repercussions. In all these scenarios, a dialogue agent must identify the safety concerns in the given setting, recognize when an interlocutor is in a vulnerable position, monitor their condition, and act accordingly. This involves a complex interplay of systems, where performance guarantees, proper objective specification, and model interpretability can be difficult. Nonetheless, these are a requirement for deployment of dialogue systems in live settings.

Absolute guarantees on safety will most likely be impractical or impossible to produce, as it would require a complete understanding of the definition of safety in dialogue. However, it should be possible to produce \textit{conditional} safety guarantees given a model trained to predict biased or unsafe behaviour in a dialogue system. In this case, one could put an upper bound on the probability that a dialogue system generates an unsafe response with respect to the model. To this end, further investigation of the aforementioned topics is needed in the context of dialogue agents: (1) risk identification and model-based assessment; (2) performance and safety guarantees in models; (3) interpretability, interruptability, and human intervention. In systems that can learn in real-time -- as in reinforcement learning (RL) -- these considerations become more complex as model performance may not be assessed and validated in an offline fashion.


\section{Reinforcement Learning (RL)}

As RL becomes more commonplace in dialogue systems~\cite{li2016deep,serban2017deep}, special considerations are needed for such agents according to the previous discussions in safety. Since RL agents learn from interaction with their environment, performance guarantees and objective formulation have slightly different concerns. Performance guarantees in this context comprise the field of safe RL~\cite{garcia2015comprehensive}. The main aspects of safe RL include modification of the optimality criterion with a safety factor or the exploration process with extrinsic information or a risk analysis metric. The latter is a concern unique to RL since agents balance exploration and exploitation during the learning process. Due to the sensitive nature of some contexts, placing guarantees on the exploratory space -- either through theoretical guarantees on the added exploration policy or through structural restriction on the exploration space -- is extremely important. If the agent uses exploratory behaviours in a live setting, this could be fatal (e.g. as in the critical settings discussed previously). Similarly, formulation of a proper objective in RL correlates to: generating a reward function which discourages and places bounds on unsafe behaviour; restricting the policy to actions and states which guarantee privacy, safety, and lack of bias. As with any optimization function, improper specification of a reward function or policy space can be catastrophic, especially in agents that learn in real time as they interact with users~\cite{neff2016automation}. However, evaluation of dialogue is still unsolved and identification of safety risks can involve separate complex mechanisms. As such, further investigation is needed to both understand dialogue safety and evaluation in order to place such guarantees on RL policies, reward functions, and objective formulations.


\section{Reproducibility}
Special consideration must also be placed on the reproducibility of dialogue systems research. Modern training of complex dialogue systems often requires specialized training schedules~\cite{li2016deep} that are difficult to replicate. Investigations of reproducibility in dialogue systems research can and should comprise singular works, akin to~\cite{hendersonRL2017}. Here, we simply emphasize that, to ensure ethical dialogue system research practices, it is paramount to release code, pre-trained models, and intricate details used to train data-driven models. Since evaluation of dialogue is an open problem, supplementary qualitative evaluation through interaction with the model may be necessary for insights into true performance, yet this is only possible if the model and detailed procedures are released. Without this it is easy to provide unfair comparisons through cherry-picked evaluations. Overall, further effort is needed to find reproducible, ethical, and fair evaluation methodologies.

\section{Discussion}
Here, this paper highlights several aspects of safety and ethics in developing dialogue systems: bias, adversarial examples, privacy, safety, considerations for RL, and reproducibility. We give demonstrative examples of ethical or safety issues in these contexts and provide recommendations based on literary and experimental findings for each direction of ethics and safety in conversational models. Our goal is to spur discussion and lines of technical data-driven dialogue systems research, including: battling underlying bias and adversarial examples; ensuring privacy, safety, and reproducibility. We hope that dialogue systems of the future can place guarantees to make obsolete the issues we discuss here, and ensure ethical and safe human-like interfaces.  






\fontsize{9.0pt}{10.0pt}\selectfont
\bibliographystyle{aaai}
\bibliography{aaai}

\end{document}